\documentclass[11pt]{article}

\usepackage[margin=1in]{geometry}
\usepackage{graphicx}
\usepackage{booktabs}
\usepackage{tabularx}
\usepackage{longtable}
\usepackage{amsmath, amssymb}

\usepackage[T1]{fontenc}
\usepackage{lmodern}          
\usepackage{microtype}        
\usepackage{setspace}
\usepackage{enumitem}
\setlist{nosep,leftmargin=*}

\usepackage{titlesec}
\titlespacing*{\section}{0pt}{1.2ex plus 0.2ex minus 0.2ex}{0.8ex}
\titlespacing*{\subsection}{0pt}{1.0ex plus 0.2ex minus 0.2ex}{0.6ex}
\titlespacing*{\subsubsection}{0pt}{0.8ex plus 0.2ex minus 0.2ex}{0.4ex}

\usepackage{xcolor}
\usepackage{hyperref}
\hypersetup{
  colorlinks=true,
  linkcolor=blue,
  citecolor=blue,
  urlcolor=blue
}

\usepackage[font=small,labelfont=bf]{caption}

\usepackage{float}

\renewenvironment{abstract}
{\small
\begin{center}\bfseries ABSTRACT\end{center}
\noindent\ignorespaces}
{\par}

\title{\Large\bfseries
Decoding the Human Factor:\\
High Fidelity Behavioral Prediction for Strategic Foresight}

\author{\normalsize
\textbf{OMGene AI Lab}\\
Ben Yellin PhD, Ehud Ezra, Mark Foreman PhD, Shula Grinapol PhD}

\date{\normalsize Version: v1.0 \;|\; January 19, 2026}

\begin{document}
\maketitle

\begin{abstract}
Predicting human decision-making in high-stakes environments remains a central challenge for artificial intelligence. While
large language models (LLMs) demonstrate strong general reasoning, they often struggle to generate consistent, individual-specific
behavior, particularly when accurate prediction depends on complex interactions between psychological traits and situational
constraints. Prompting-based approaches can be brittle in this setting, exhibiting identity drift and limited ability to leverage
increasingly detailed persona descriptions.

To address these limitations, we introduce the \emph{Large Behavioral Model (LBM)}, a behavioral foundation model fine-tuned to
predict individual strategic choices with high fidelity. LBM shifts from transient persona prompting to behavioral embedding by
conditioning on a structured, high-dimensional trait profile derived from a comprehensive psychometric battery. Trained on a
proprietary dataset linking stable dispositions, motivational states, and situational constraints to observed choices, LBM learns
to map rich psychological profiles to discrete actions across diverse strategic dilemmas.

In a held-out scenario evaluation, LBM fine-tuning improves behavioral prediction relative to the unadapted
\texttt{Llama-3.1-8B-Instruct} backbone and performs comparably to frontier baselines when conditioned on Big Five traits. Moreover,
we find that while prompting-based baselines exhibit a complexity ceiling, LBM continues to benefit from increasingly dense trait
profiles, with performance improving as additional trait dimensions are provided. Together, these results establish LBM as a
scalable approach for high-fidelity behavioral simulation, enabling applications in strategic foresight, negotiation analysis,
cognitive security, and decision support.
\end{abstract}

\section{Introduction}
Explaining and predicting human choice under risk and uncertainty has long been a central challenge in the behavioral sciences,
from early utility-based theories to modern accounts of systematic deviations from rational choice
(Bernoulli, 1954; Kahneman \& Tversky, 1979; Tversky \& Kahneman, 1992).
In high-stakes environments - including conflict escalation, resource negotiation, and strategic adaptation - decisions are often
shaped by stable dispositions and cognitive constraints rather than chance or general-purpose reasoning.
While large language models (LLMs) have achieved strong performance in language understanding and general reasoning, recent evidence
suggests they remain limited as behavioral simulators of specific individuals, particularly when accurate prediction depends on
complex and interacting psychological traits (Hu et al., 2025; Scientific Reports, 2025; NeurIPS, 2024).

A core difficulty is that human decision-making is inherently high-dimensional. Even in narrowly defined strategic dilemmas, a
single choice may reflect the interaction of multiple psychological factors - such as threat sensitivity, impulse control,
long-term planning, and moral disengagement - whose effects depend on situational context and on each other. Despite the importance
of these interactions, there remains limited high-quality evidence linking rich, multi-trait profiles to observed decision
outcomes at the individual level, especially in settings where behavior unfolds strategically over time. This gap constrains both
theory-driven and data-driven approaches, which often rely on coarse trait summaries, simplified tasks, or population-level
averages that obscure meaningful individual differences.

One widely used approach is to treat LLMs as ``virtual participants'' through persona prompting, where the model is conditioned on
a natural-language description of identity, traits, or preferences. However, prior work suggests that standard LLMs do not
reliably seek, store, or update missing person-specific information, leading to inconsistent behavior over longer interactions
(Tint et al., 2024; Kwan et al., 2024). Related studies also report identity drift, where models gradually deviate from an assigned
persona as interactions extend (Choi et al., 2024). These issues are exacerbated in long-context settings, where transformer
models can underutilize information placed in the middle of long inputs (``lost in the middle'')
(Liu et al., 2023; Hsieh et al., 2024). In practice, this creates a context bottleneck: only a fraction of a long persona
description meaningfully influences the model’s predictions, limiting fidelity to individual-specific behavioral tendencies.

A second line of work moves beyond prompting and instead trains behavioral foundation models directly on large-scale human decision
datasets. Recent efforts demonstrate that fine-tuning on behavioral responses can improve simulation accuracy compared to
prompt-only baselines. For example, Centaur was introduced as a foundation model of human cognition trained on large behavioral
datasets (Binz et al., 2025), and SOCRATES reported substantial gains in behavioral simulation using a large dataset of human
responses (Kolluri et al., 2025). Additional work such as Be.FM reports improvements in personality prediction and scenario-based
behavior simulation relative to standard reasoning models (University of Michigan, 2025). At the same time, benchmark evidence
suggests that even when LLMs capture broad qualitative trends, they can still differ systematically from human behavior
distributions and behave inconsistently in interactive settings (Hu et al., 2025; Scientific Reports, 2025; NeurIPS, 2024).

Motivated by these limitations, we introduce the Large Behavioral Model (LBM), which shifts behavioral conditioning from
natural-language persona prompts into the model’s internal representations. Rather than encoding high-dimensional personas solely
as text, LBM incorporates structured trait information as a persistent conditioning signal, with the goal of improving stability,
reducing identity drift, and enabling more consistent simulation of individual-specific decision tendencies across diverse
strategic scenarios. LBM is developed and evaluated using a proprietary dataset that links psychometric profiles to scenario-based
strategic choices, enabling systematic evaluation of generalization to unseen situational contexts.

In summary, this work makes three contributions. First, we introduce a fine-tuned behavioral foundation model that conditions on
structured, high-dimensional trait profiles for individual-level behavioral prediction. Second, we present a large-scale dataset
linking psychometric measures to strategic decision outcomes across diverse scenarios. Third, we provide empirical evidence that
fine-tuning enables performance improvements over prompting-based baselines, and that model performance scales with trait
dimensionality beyond what prompting alone can effectively leverage.

\section{Large Behavioral Model (LBM)}
\subsection{Task Definition}
We formulate behavioral prediction as a supervised learning problem in which a model predicts how a
specific individual will act in a strategic dilemma, conditioned on both the scenario context and a
stable participant profile. For a participant $u$ and a scenario $s$, the model learns
\begin{equation}
P(y \mid s, t_u),
\end{equation}
where $t_u \in \mathbb{R}^{K}$ is a high-dimensional trait representation derived from standardized
psychological measures, and $y$ denotes the observed behavioral outcome (i.e., a discrete multiple-choice
response corresponding to a strategic action).

Each training example consists of the tuple $(t_u, s)$, where $t_u$ encodes the participant profile and
$s$ is a natural-language scenario prompt describing the strategic dilemma. Scenarios are designed to elicit
meaningful variation in decision tendencies and include explicit constraints such as stakes (e.g., financial,
reputational, or relational consequences), ambiguity (incomplete information), and urgency (time pressure).
The model outputs a discrete behavioral prediction $\hat{y}$ (e.g., escalate, withdraw, compromise) and may
optionally produce a bounded rationale trace $e$, enabling deterministic parsing and downstream integration.

\subsection{Data Collection and Participants}
Participant data are collected through the OMGene application. Recruitment is conducted primarily via
self-registration through the study website \url{https://app.omgene.ai/register}, complemented by volunteer outreach
and snowball sampling across multiple cohorts (e.g., in-person recruitment, close-circle outreach,
existing/new users, students, and external recruitment channels). The protocol targets a planned sample
size of 2{,}500 participants to ensure sufficient statistical power for high-dimensional analysis.

\begin{table}[t]
\centering
\caption{Dataset Specifications and Study Design}
\label{tab:dataset_specs}
\begin{tabular}{lll}
\hline
\textbf{Component} & \textbf{Metric / Scope} & \textbf{Notes} \\
\hline
Sample Size & 2{,}500 participants & Enrollment (Jan 2024 -- Dec 2025) \\
Recruitment Method & Volunteer / Snowball & Self-registration via OMGene App \\
Psychometric Data & Comprehensive Battery & Administered to all valid enrollees \\
Behavioral Data & Scenario-Behavior Dataset & Sourced from a bank of 55 unique scenarios \\
Data Coverage Strategy & Multi-Scenario Response & Maximizes observed vs. possible interactions \\
Population A & English Speakers (USA) & Primary cohort \\
Population B & Hebrew Speakers (Israel) & Secondary cohort \\
\hline
\end{tabular}
\end{table}

The dataset integrates a comprehensive psychometric battery alongside scenario-based behavioral data.
The behavioral component draws from a bank of 55 unique scenarios. Participants engage with multiple
scenarios (but not necessarily the full set), yielding a partially observed participant--scenario matrix.
This data coverage strategy increases the number of observed responses relative to possible participant--scenario
interactions (Table~\ref{tab:dataset_specs}).

Eligibility criteria include adults aged 18--99, written informed consent, and willingness/ability to adhere
to study procedures. Exclusion criteria include prominent features of mania/psychosis and/or cognitive
impairment, as well as self-reported suicidal ideation. Participants may withdraw at any time and may request
deletion of their collected data, subject to consent-defined retention and applicable legal requirements.

In addition to psychometric and scenario data, we collect socio-demographic and contextual variables for
dataset characterization, including age, gender, familial status, education, occupation, socioeconomic status
(SES), employment status, ethnicity, residential area, access to services, time-related resources, and meaningful
or stressful life events. Recruitment spans January 2024 to December 2025. Because recruitment relies on
volunteer self-registration rather than probability sampling, the cohort is best characterized as a convenience
volunteer sample. The target population composition includes a predominantly USA-based English-speaking cohort,
alongside a smaller Israeli Hebrew-speaking sub-cohort.

\paragraph{Ethics, Consent, and Data Governance.}
All participants receive an explanation of the study objectives and procedures and provide written informed
consent prior to inclusion. Participants showing elevated risk indicators during intake (e.g., prominent paranoid
thinking, psychosis risk, or suicidal ideation) are excluded from participation and are advised to seek evaluation
through community mental health services. If participants experience distress during questionnaires or system
interaction, immediate support is offered by the PI or an appointed clinical expert.

The study is conducted under institutional Helsinki Committee procedures and follows Ministry of Health and Good
Clinical Practice (GCP) requirements. The protocol references Helsinki approval number 05/25. Data are processed
under coded identifiers (pseudonymized), with direct identifiers excluded from research datasets and stored separately
under restricted access. Where participants provide free-text responses or uploads that may contain identifying
information, an operational process removes identifying information within 30 days. Questionnaire data are retained
for 7 years in accordance with the protocol.

\subsection{Psychological Measures}
Participants’ psychological profiles are computed from multiple validated psychometric instruments and
aggregated into a unified representation designed for individual-level behavioral prediction. We adopt an
``architecture of action'' framing, where behavior reflects who the participant is, the situation they face,
and the processes that translate a person-in-situation into action (Beck \& Jackson, 2022; Ajzen, 1991).
Accordingly, the profile integrates complementary components rather than relying on a single construct,
including (i) stable dispositions, (ii) motivational and goal processes, (iii) self-regulation and control
capacity, (iv) affective and stress-related processes, and (v) social and structural context. In addition,
because repeated experiences can produce automatic cue--response patterns, the profile is designed to support
modeling of history-driven action tendencies, consistent with habit-based accounts of behavior
(Wood \& Neal, 2007).

Instruments are scored using standard procedures (including reverse-coded items where applicable) and
aggregated into a unified trait representation ($K = 74$ traits in the current configuration; full instrument
list and mapping in Appendix~A). Trait values are standardized and discretized into ordinal bins to improve
interpretability and to better support consistent conditioning in the model input.

\subsection{Scenarios and Behavioral Responses}
LBM is trained on a scenario bank designed to elicit decision-making in strategic contexts characterized
by conflict, trust, and uncertainty. These settings are chosen to capture meaningful variation in how
individuals respond when incentives shift, agreements are tested, and reputational or relational consequences
are salient. The scenario bank contains 55 distinct scenarios in total. Each participant answers multiple
scenarios, but not necessarily the full set, resulting in a partially observed participant--scenario matrix
that supports learning both within-person consistency and between-person variation.

Scenarios are organized into three types: Major Life Events (retrospective), Day-to-Day Events (last 12 months),
and Hypothetical Scenarios. The set spans six behavioral domains: Trust Dynamics, Conflict and Resolution,
Power and Influence, Risk and Decision Behavior, Integrity and Compliance, and Strategic Adaptation. Across
scenario types, prompts are written to vary systematically along key dimensions such as stakes, ambiguity,
and urgency.

Each scenario contains multiple question formats to support supervised learning and analysis at different
levels of structure. Participants provide open-text responses, multiple-choice responses that map to discrete
behavioral labels $y$, and Likert-scale ratings (1--5) that quantify internal state variables. This mixed-format
design enables LBM to predict structured behavioral outcomes while optionally producing bounded rationale traces
aligned with the participant profile and scenario constraints.

\subsection{Configuration and Training}
LBM is a large language model adapted for individual-level behavioral prediction in strategic scenarios.
Standard persona prompting typically encodes participant traits as free-form natural language, which can be
brittle under long contexts and may drift toward generic or socially average responses. LBM instead conditions
generation on a structured participant profile that is injected consistently into every scenario prompt.

\subsubsection{Input representation: structured profile and scenario prompts}
LBM builds on Llama-3.1-8B-Instruct as the instruction-following backbone. Each training and inference example
is formatted as a deterministic prompt that combines:
(i) a scenario type identifier (e.g., day-to-day, retrospective, or hypothetical),
(ii) a participant profile block containing demographic attributes (e.g., age, sex) and a structured list of
psychological trait values derived from the psychometric battery,
(iii) a natural-language scenario description,
(iv) optional structured context fields consisting of follow-up questions and participant-provided background, and
(v) one or more multiple-choice prediction questions with answer options.
The model is instructed to predict the participant’s answer for each question and return only a single-line JSON
object mapping question identifiers to option numbers.

\subsubsection{Output format}
For multi-question scenarios, the model outputs a JSON object containing (i) predicted option indices for each
question and (ii) an optional bounded rationale trace aligned to each prediction:
\begin{verbatim}
{
"predictions": {"Q4": 5, "Q5": 3},
"reasoning": {"Q4": "...", "Q5": "..."}
}
\end{verbatim}

\subsubsection{Supervised fine-tuning objective}
LBM is trained via supervised fine-tuning on participant--scenario examples. Each training sample pairs a
serialized participant profile and scenario prompt with a target output containing the ground-truth behavioral
label(s) $y$, and, when available, a bounded rationale trace $e$. Training optimizes a weighted likelihood objective over the structured output, emphasizing correct prediction
of the discrete answer choice for each question.

\subsubsection{Parameter-efficient adaptation (LoRA)}
To preserve the backbone model’s general capabilities while specializing it for behavioral prediction, LBM uses
parameter-efficient fine-tuning with LoRA. We apply LoRA adapters to all linear layers. We use LoRA rank ($r = 16$),
scaling ($\alpha = 32$), dropout 0.1, and enable RS-LoRA.

\subsubsection{Optimization and inference}
Training is performed for 2 epochs with learning rate $5 \times 10^{-5}$, warmup, and gradient clipping for stability.
At inference time, the model is constrained to output valid JSON according to the schema above, and a lightweight
parsing layer extracts predictions deterministically.

\section{Results}
We evaluate the Large Behavioral Model (LBM) using a held-out scenario split to assess
generalization to unseen situational contexts. Specifically, we train on 75\% of scenarios and
evaluate on the remaining 25\% held-out scenarios.

Because the response labels are class-imbalanced, we report balanced accuracy and macro-F1 as
primary metrics, as they better reflect performance across classes. We additionally report standard
accuracy as a familiar reference measure. Error bars denote 95\% confidence intervals computed via
participant-level bootstrap resampling. Finally, since most questions correspond to a 5-way
multiple-choice prediction task, chance performance is approximately 0.20.

\subsection{LBM fine-tuning improves behavioral prediction and performs comparably to frontier baselines under Big Five trait conditioning}

Across held-out evaluation scenarios in the Big Five setting, LBM fine-tuning improves behavioral
prediction performance over the Llama base model across all metrics. Specifically, the SFT-tuned
LBM increases accuracy from 0.42 to 0.48, balanced accuracy from 0.24 to 0.31, and macro-F1 from
0.16 to 0.26, indicating consistent gains when conditioning only on the Big Five personality traits.

Among the frontier LLM baselines evaluated, Claude 4.5 Sonnet is the strongest performer, achieving
an accuracy of 0.463, balanced accuracy of 0.307, and macro-F1 of 0.250. Claude outperforms other
frontier baselines - including GPT-5 Mini, GPT-4o Mini, Grok 4, and DeepSeek V3 - across the three
metrics in this setting. At the same time, LBM performs comparably to Claude, with differences
falling within the 95\% confidence intervals, suggesting that their performance is not reliably
distinguishable given statistical uncertainty.
Complete summary statistics for Figure~1 are reported in Appendix~C (Table~\ref{tab:fig1_results}).

Overall, these results show that LBM fine-tuning yields significant improvements over the (\texttt{Llama-3.1-8B-Instruct} base
model, and that LBM and Claude 4.5 Sonnet are the strongest-performing models in the Big Five setting.

\begin{figure}[tbp]
\centering
\includegraphics[width=\textwidth]{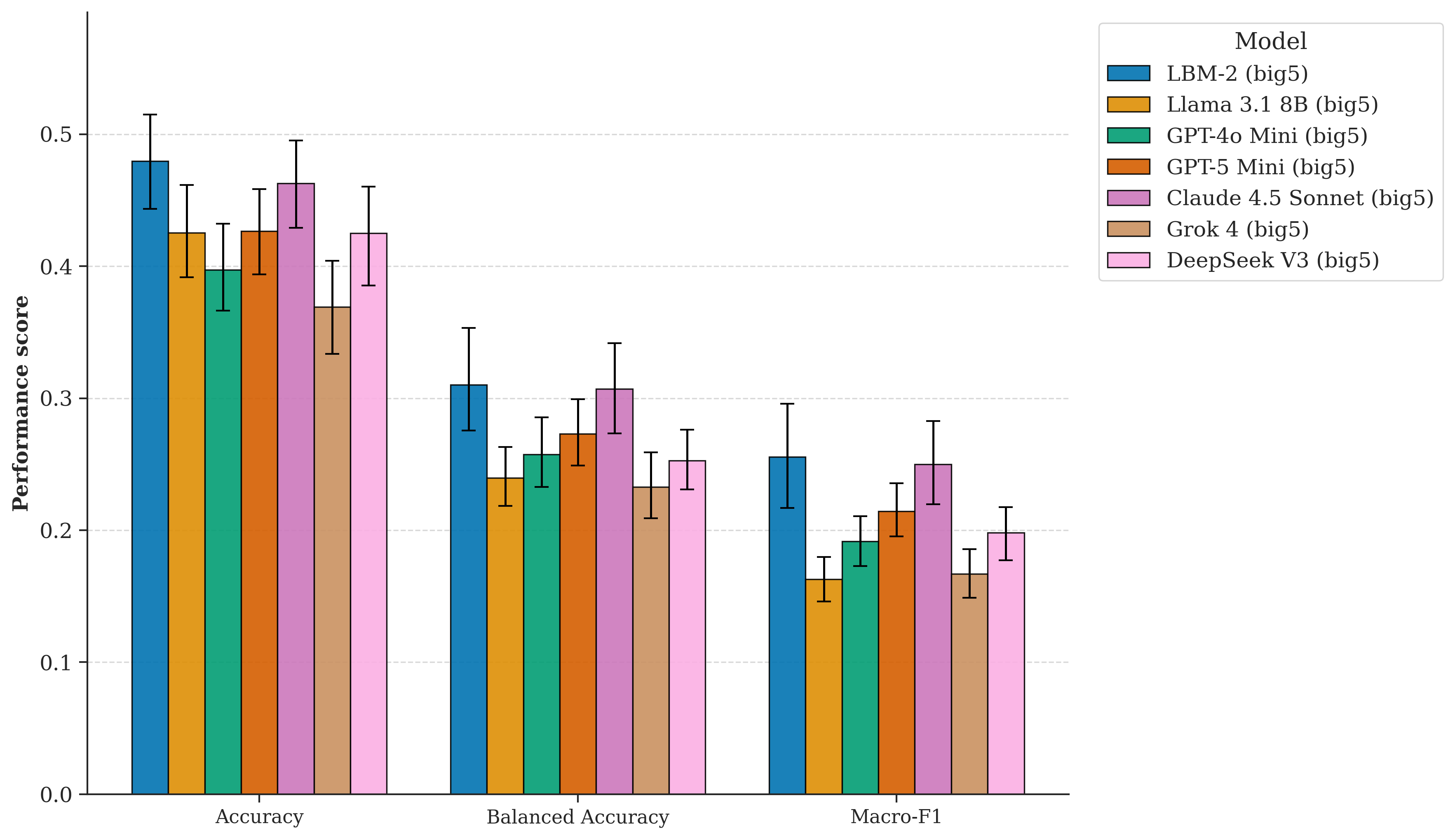}
\caption{Comparison of Behavioral Alignment across LLMs. Bars report accuracy, balanced accuracy, and macro-F1 for the base backbone model, the SFT-tuned LBM, and frontier LLM baselines. Error bars indicate 95\% confidence intervals computed using participant-level bootstrap resampling. (Rendered from PDF page 11.)}
\label{fig:llm_comparison}
\end{figure}

\subsection{Scaling trait dimensionality improves LBM performance, but not prompting-based baselines}
We compare the fine-tuned LBM against three baselines: the untuned backbone model
(\texttt{Llama-3.1\allowbreak-8B\allowbreak-Instruct}),
Claude 4.5 Sonnet, and Grok 4. For each model, we vary the size of the trait profile provided in the
prompt, using 5, 10, 20, 40, and 74 traits, while keeping the evaluation protocol fixed.

LBM consistently benefits from richer trait conditioning. As the number of traits increases from 5 to 10
and then to 20, LBM performance improves from 0.48 to 0.56 to 0.62 in accuracy, with corresponding gains
in macro-F1 from 0.26 to 0.39 to 0.45. Beyond 20 traits, performance exhibits diminishing returns, without
any additional gains when expanding the profile to 40 and 74 traits 

In contrast, the prompt-based baselines exhibit limited sensitivity to additional trait information. Both
Claude and the base (\texttt{Llama-3.1-8B-Instruct} model remain approximately stable across trait dimensionality, suggesting that
prompting alone may not effectively leverage increasingly dense psychological profiles. Grok exhibits a
modest upward trend as trait dimensionality increases toward 40 traits, but remains below Claude overall
and does not match the improvements achieved by the fine-tuned LBM. Together, these results support the
existence of a prompting ``complexity ceiling,'' while demonstrating that fine-tuning enables more
effective integration of high-dimensional trait information for behavioral prediction. Complete summary statistics for Figure~2 are reported in Appendix~C (Table~\ref{tab:fig2_results}).

\begin{figure}[tbp]
\centering
\includegraphics[width=\textwidth]{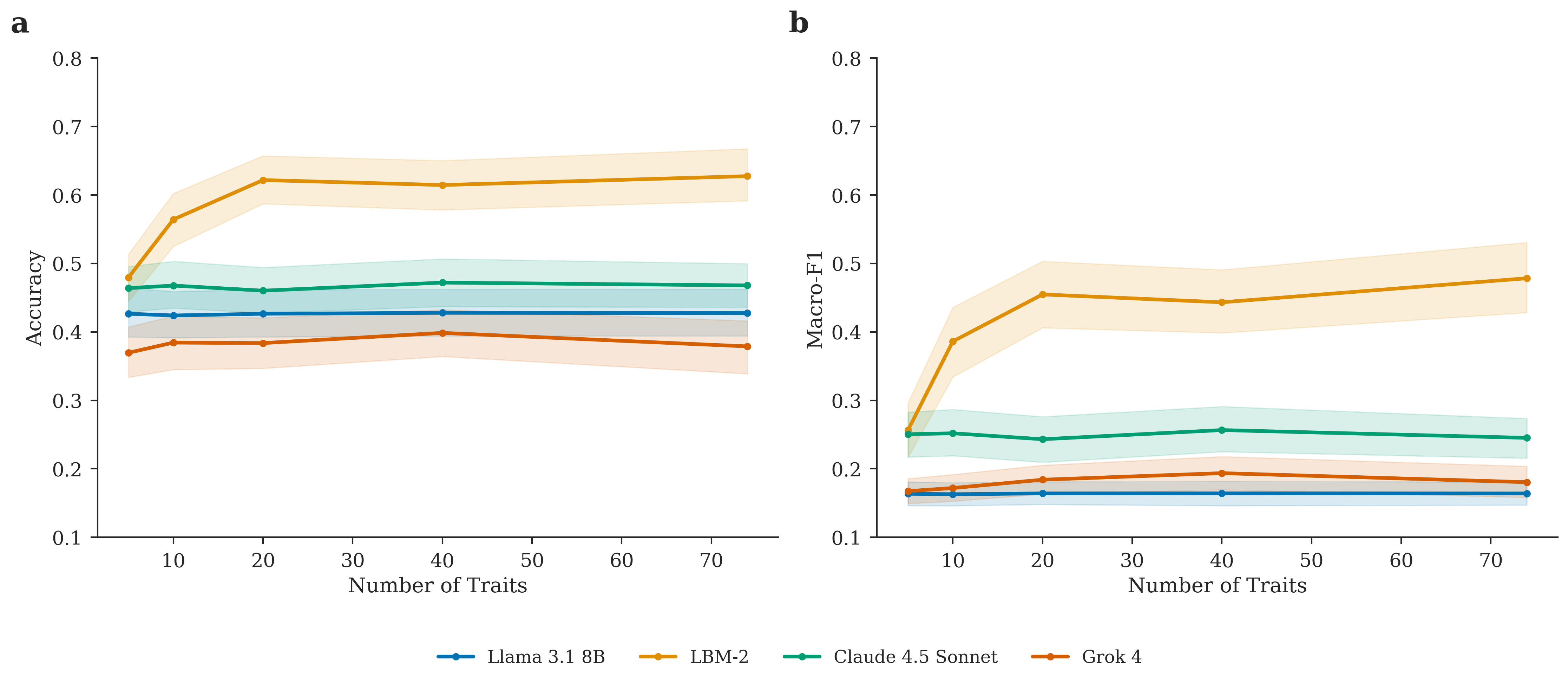}
\caption{Effect of trait dimensionality on behavioral prediction. Lines show model performance when conditioned on 5, 10, 20, 40    and 74 traits. (a) Accuracy. (b) Macro-F1. LBM improves with additional traits, while the base model and Claude remain approximately constant; Grok shows a modest upward trend but remains below LBM.}
\label{fig:trait_scaling}
\end{figure}

\section{Discussion}
The Large Behavioral Model (LBM) represents a shift from persona prompting, where identity is treated as transient context,
to behavioral embedding, where individual dispositions are integrated as a stable conditioning signal. Across held-out
evaluation scenarios, our results show that parameter-efficient fine-tuning on high-quality psychometric and scenario-based
data improves the ability of large language models to predict individual strategic choices.

\subsection{Superiority of Fine-Tuning over Contextual Prompting}
A central finding is that the fine-tuned LBM outperforms the unadapted \texttt{Llama-3.1-8B-Instruct} backbone in behavioral
prediction, and performs comparably to strong frontier baselines under Big Five trait conditioning. This gap is consistent
with limitations of purely prompt-based approaches, which rely on the model’s ability to reliably attend to and utilize
long, high-dimensional persona descriptions at inference time. In contrast, fine-tuning enables the model to internalize
systematic mappings from trait profiles and situational cues to action distributions, reducing reliance on brittle
context-only conditioning.

\subsection{The Role of Trait Dimensionality and Saturation}
LBM performance improves as the trait profile expands from 5 to 20 traits, indicating that the model can effectively
integrate richer psychometric information to refine its behavioral predictions. Beyond this range, performance exhibits
diminishing returns under the current dataset scale and scenario coverage. In contrast, prompt-based baselines show
limited sensitivity to additional trait information, consistent with a complexity ceiling in which increasingly dense
persona conditioning does not translate into improved behavioral prediction. Together, these results suggest that
fine-tuning is necessary to reliably leverage high-dimensional psychological profiles, particularly when prediction depends
on interactions among multiple traits.

\subsection{Limitations}
Our study has several limitations. First, the current cohort is a convenience volunteer sample, predominantly USA-based
English speakers with a smaller Israeli Hebrew-speaking sub-cohort, which may limit generalizability to other populations.
Second, the dataset relies in part on retrospective recall and hypothetical scenarios, and an intention--behavior gap may
remain between reported decisions and real-world actions. Third, while the scenario prompts are designed to vary stakes,
ambiguity, and urgency, textual descriptions cannot fully replicate the sensory and emotional immersion of live interactions.
Future work should evaluate behavioral prediction under more ecologically valid settings and with broader population coverage.

\subsection{Conclusion and Future Directions}
This work provides evidence that high-fidelity behavioral simulation benefits from models structurally adapted to the
architecture of action. Future development will focus on scaling scenario complexity, expanding the diversity of participant
profiles, and modeling longer-horizon interactions that capture dynamic social feedback. A key objective is to maintain high
identity consistency while improving the model’s ability to simulate increasingly complex strategic behavior across contexts.

\clearpage
\section*{Citations}
\begin{enumerate}
\item Bernoulli, D. (1954). \textit{Exposition of a New Theory on the Measurement of Risk} (translation of 1738 work). \textit{Econometrica}, 22, 23--36.
\item Kahneman, D., \& Tversky, A. (1979). Prospect Theory: An Analysis of Decision under Risk. \textit{Econometrica}, 47, 263--292.
\item Tversky, A., \& Kahneman, D. (1992). Advances in Prospect Theory: Cumulative Representation of Uncertainty. \textit{Journal of Risk and Uncertainty}, 5, 297--323.
\item Hu, T., Baumann, J., Lupo, L., Collier, N., Hovy, D., \& R\"ottger, P. (2025). SimBench: Benchmarking the Ability of Large Language Models to Simulate Human Behaviors. arXiv.
\item Scientific Reports (Nature Portfolio). (2025). Evaluating the ability of large language models to predict human social decision outcomes (title may vary slightly by indexing). \textit{Scientific Reports}.
\item NeurIPS (Proceedings). (2024). Can Large Language Model Agents Simulate Human Trust Behavior? NeurIPS 2024.
\item Tint et al. (2024). [Your exact paper title here].
\item Kwan et al. (2024). [Your exact paper title here].
\item Choi, J., Hong, Y., Kim, M., \& Kim, B. (2024). Examining Identity Drift in Conversations of LLM Agents. arXiv.
\item Liu, N. F., Lin, K., Hewitt, J., Paranjape, B., Bevilacqua, M., Liang, P., \& Hashimoto, T. B. (2023). Lost in the Middle: How Language Models Use Long Contexts. arXiv.
\item Hsieh, C.-Y., et al. (2024). Found in the Middle: Calibrating Positional Attention Bias Improves Long Context Utilization. arXiv.
\item Binz et al. (2025). Centaur: [full title]. \textit{Nature} (as cited).
\item Kolluri et al. (2025). SOCRATES: [full title]. EMNLP 2025 (as cited).
\item University of Michigan News / Be.FM (2025). AI that thinks like us: U-M researchers unveil new model to predict human behavior. (News release describing Be.FM).
\item CoSER (2025). Complex Persona Simulation \& Evaluation. [venue/link as cited].
\end{enumerate}

\clearpage
\appendix
\section{Psychological Measures}
\subsection{Instrument Battery}
The following instruments are administered. Each is scored using the published scoring key and
then mapped into the unified trait representation used by LBM.

\paragraph{Core dispositions / interpersonal tendencies}
\begin{itemize}
\item NEO-FFI (Big Five)  -  A 60-item self-report questionnaire measuring Neuroticism, Extraversion, Openness, Agreeableness, and Conscientiousness (5-point Likert scale).
\item EPQ Lie Scale (EPQ-L)  -  A 12-item validity subscale from the Eysenck Personality Questionnaire Revised (EPQ-R) designed to detect social desirability or potential response distortion.
\item Brief-COPE  -  A 16-item inventory assessing coping strategies (adaptive and maladaptive).
\item Multifactor Leadership Questionnaire (MLQ - Form 5X)  -  A 36-item scale assessing leadership styles and outcomes.
\item UPPS-P Impulsive Behavior Scale  -  A 59-item instrument measuring five facets of impulsivity.
\item Barratt Impulsiveness Scale (BIS-11)  -  A 30-item self-report questionnaire measuring trait impulsivity.
\item Intolerance of Uncertainty Scale (IUS-12)  -  A 12-item measure evaluating distress in response to uncertainty.
\item Optimism Scale (LOT)  -  The Life Orientation Test, an 8-item scale measuring dispositional optimism.
\item Hope Scale  -  A 12-item instrument assessing goal-directed thinking (agency and pathways).
\item Mastery Scale  -  A 7-item scale measuring sense of personal control over life events.
\item Connor-Davidson Resilience Scale (CD-RISC-10)  -  A 10-item scale assessing resilience.
\item Emotion Regulation Questionnaire (ERQ)  -  A 10-item scale measuring cognitive reappraisal and expressive suppression.
\item Parental Bonding Instrument (PBI)  -  A 25-item questionnaire assessing perceptions of parental care and overprotection.
\item Multidimensional Scale of Perceived Social Support (MSPSS)  -  A 12-item scale evaluating perceived social support.
\item DeJong Gierveld Loneliness Scale  -  A 6-item scale measuring emotional and social loneliness.
\item WHO-5 Well-Being Index  -  A 5-item screening tool for subjective well-being.
\item Brief Symptom Inventory (BSI)  -  A 53-item self-report inventory assessing psychological distress.
\item PCL-5 (PTSD Checklist)  -  A 20-item self-report tool for PTSD symptom severity (DSM-5).
\item ADHD Adult \& Adolescent Screening  -  An 18-item scale evaluating inattention and hyperactivity/impulsivity (DSM-5).
\item Alcohol Use Disorders Identification Test (AUDIT)  -  A 10-item screening tool for hazardous alcohol use.
\end{itemize}

\subsection{Trait table (First 15 Traits)}
\setlength{\LTpre}{0pt}
\setlength{\LTpost}{0pt}
\begin{longtable}{@{}p{0.17\textwidth}p{0.20\textwidth}p{0.12\textwidth}p{0.12\textwidth}p{0.33\textwidth}@{}}
\caption{Trait table (first 15 traits shown in PDF)}\label{tab:traits15}\\
\toprule
Trait & Short description & Instrument & Items & Standardization and binning \\
\midrule
\endfirsthead
\toprule
Trait & Short description & Instrument & Items & Standardization and binning \\
\midrule
\endhead
Openness & Curiosity, imagination, aesthetic and intellectual openness & NEO-FFI & 12 &
S: Gen pop (Schmitt et al., 2007). B: Very low ($z<-2\sigma$), Low ($-2\sigma<z<-\sigma$), Normal ($-\sigma<z<\sigma$), High ($\sigma<z<2\sigma$), Very high ($z>2\sigma$). \\
Conscientiousness & Organization, dutifulness, persistence, and self-discipline & NEO-FFI & 12 &
S: Gen pop (Schmitt et al., 2007). B: Same as Openness. \\
Extraversion & Sociability, assertiveness, energy, and positive emotionality & NEO-FFI & 12 &
S: Gen pop (Schmitt et al., 2007). B: Same as Openness. \\
Agreeableness & Trust, altruism, compliance, and modesty & NEO-FFI & 12 &
S: Gen pop (Schmitt et al., 2007). B: Same as Openness. \\
Neuroticism & Tendency to experience negative emotions like anxiety and anger & NEO-FFI & 12 &
S: Gen pop (Schmitt et al., 2007). B: Same as Openness. \\
Lie / Social Desirability & Social desirability, overly virtuous self-presentation, potential response distortion & EPQ-RS & 12 &
S: None. B: (0--4) Low/realistic; (5--8) Normal; (9--12) High socially desirable reporting. \\
Impulsivity & Tendency to act on impulse across attentional, motor, and non-planning domains & BIS-11 & 30 &
S: None. B: $\le 52$ Low; 53--71 Average; $\ge 72$ High. \\
Resilience & Ability to cope with stress and bounce back from adversity & CD-RISC-10 & 10 &
S: General population norms. B: Quartile based (Low, Normal, High). \\
Optimism & Generalized expectation of positive outcomes & LOT & 8 &
S: None. B: Tertile split (Low, Moderate, High) based on raw score (8--40). \\
Intolerance of Uncertainty & Cognitive and behavioral distress in response to uncertainty & IUS-12 & 12 &
S: Clinical / Non-clinical norms. B: Low (12--30), Moderate (31--45), High (46--60). \\
Social Support & Perceived support from family, friends, and significant others & MSPSS & 12 &
S: None. B: Low (1--2.9), Moderate (3--5), High ($>5$) based on mean score. \\
Coping Style (Adaptive) & Use of active coping, planning, and positive reframing strategies & Brief-COPE & 12 &
S: None. B: Median split or tertiles (Low, Moderate, High usage). \\
Psychological Distress (GSI) & Overall level of current psychological symptom severity & BSI & 53 &
S: Clinical / community norms. B: T-scores; $\ge 63$ indicates clinical significance (High). \\
Transformational Leadership & Proactive, inspirational, and influence-based leadership style & MLQ & 20 &
S: Leadership norms. B: Low (0--1.9), Moderate (2.0--2.9), High (3.0--4.0). \\
PTSD Severity & Severity of intrusive, avoidance, and arousal symptoms & PCL-5 & 20 &
S: Clinical cut-offs. B: $<33$ Non-clinical; $\ge 33$ Probable PTSD / high severity. \\
\bottomrule
\end{longtable}

\clearpage
\section{Prompt Examples}
\subsection*{Example 1}
\begin{verbatim}
SCENARIO_TYPE: DTD
PARTICIPANT_PROFILE:
 - age: 36.0
 - sex: male
 - traits (Z-scores):
 - Neuroticism: 0.24 (Normal)
 - Extraversion: -1.56 (Low)
 - Openness: 0.76 (Normal)
 - Agreeableness: -1.31 (Low)
 - Conscientiousness: 1.15 (High)
SCENARIO:
Think about an event from the past year in which others (colleagues, managers,
etc.) pressured you to take a shortcut or bypass rules, in a way that brings
immediate benefit to the team but violates procedures or ethical standards.
CONTEXT:
Q: Briefly describe the rule you chose to break and the benefit that was
gained. A: My immediate manager wanted us to do something that is not up to
code and would fail inspection.
Q: How strong was the social pressure to agree to the request? A: High:
Refusing would have created significant friction and made future collaboration
difficult.
Q: What was the relative cost of the action you chose for your relationship
with that person? A: Low cost: Mild, temporary discomfort occurred.
Q: How clear was it to you at the time that your action was violating ethical
standards? A: Very clear: It was a clear and unambiguous violation of
compliance rules.
TASK:
Predict the participant's immediate answer for EACH question below.
Return ONLY a single-line JSON object mapping question_id -> option_number.
QUESTIONS:
Q3:
What was the main reason for the action you chose?
OPTIONS:
1: Integrity / principle: I prioritized sticking to my personal ethical
standards.
2: Protecting the relationship: I prioritized protecting the relationship with
the person/group who asked.
3: Avoiding social consequences: I prioritized avoiding the heavy social cost
of saying “no.”
4: Cost of refusing: I prioritized avoiding the operational cost (for example,
a project delay) that refusing would have caused.
5: Personal gain: I prioritized my own benefit (for example, a bonus or reduced
workload) from the shortcut.
Q4:
What did you decide to do?
OPTIONS:
1: Full resistance: I refused to cooperate or take part.
2: Compromise: I took partial action (for example, delayed it or made a small
change to the numbers).
3: Full compliance: I did exactly what I was asked to do.
4: Escalation: I brought the issue to a neutral third party or an official
authority.
5: Disengagement: I removed myself from the situation immediately and stopped
being
\end{verbatim}

\subsection*{Example 2}
\begin{verbatim}
SCENARIO_TYPE: Retro
PARTICIPANT_PROFILE:
 - age: 22.0
 - sex: male
 - traits (Z-scores):
 - Neuroticism: 1.34 (High)
 - Extraversion: -0.71 (Normal)
 - Openness: 0.76 (Normal)
 - Agreeableness: 1.09 (High)
 - Conscientiousness: 0.35 (Normal)
SCENARIO:
Briefly describe a situation in which you faced a decision about whether to
disclose, on your own initiative, negative information about yourself.
Disclosure of the information could have damaged your image or standing, while
avoiding disclosure could have protected you in the short term or prevented
conflict.
CONTEXT:
Q: What information did you choose to conceal, and from which person or group
did you conceal it? A: During work on an important project I discovered that I
had misestimated the initial scope of the task, and the delay was partly caused
by me. I hesitated whether to inform the manager on my own initiative, because
this could have harmed my professional image, or to stay silent and hope to
close the gap by myself. Ultimately I faced a choice between transparency and
accountability and avoiding conflict in the short term.
Q: How strong was the pressure or incentive (personal, social, or financial) to
conceal the information? A: Moderate
Q: To what extent were you confident that the information would eventually be
discovered? A: To a large extent
TASK:
Predict the participant's immediate answer for EACH question below.
Return ONLY a single-line JSON object mapping question_id -> option_number.
QUESTIONS:
Q4:
What did you decide to do?
OPTIONS:
1: Voluntary disclosure: I disclosed the information immediately and accepted
the consequences.
2: Controlled disclosure: I disclosed the information strategically (for
example, minimizing exposure or revealing only part of it).
3: Active concealment: I deliberately chose not to disclose the information,
hoping it would not be discovered.
4: Active deception: I consciously fabricated
\end{verbatim}

\clearpage

\section{Summary Tables for Figures 1 and 2}

\begin{table}[H]
\centering
\caption{Figure 1 results (Big Five setting). Mean performance with 95\% participant-level bootstrap confidence intervals.}
\label{tab:fig1_results}
\small
\begin{tabular}{lllllllll}
\hline
\textbf{Model} & \textbf{Metric} & \textbf{Mean} & \textbf{Std} & \textbf{2.5\%} & \textbf{25\%} & \textbf{50\%} & \textbf{75\%} & \textbf{97.5\%} \\
\hline
LBM-2 (big5) & Accuracy & 0.480 & 0.018 & 0.443 & 0.467 & 0.479 & 0.492 & 0.515 \\
LBM-2 (big5) & Balanced Accuracy & 0.310 & 0.020 & 0.275 & 0.296 & 0.309 & 0.323 & 0.353 \\
LBM-2 (big5) & Macro-F1 & 0.256 & 0.021 & 0.217 & 0.241 & 0.256 & 0.270 & 0.296 \\
Llama 3.1 8B (big5) & Accuracy & 0.425 & 0.018 & 0.392 & 0.413 & 0.425 & 0.437 & 0.461 \\
Llama 3.1 8B (big5) & Balanced Accuracy & 0.240 & 0.011 & 0.218 & 0.232 & 0.240 & 0.247 & 0.263 \\
Llama 3.1 8B (big5) & Macro-F1 & 0.163 & 0.009 & 0.147 & 0.158 & 0.164 & 0.170 & 0.181 \\
Claude 4.5 Sonnet (big5) & Accuracy & 0.463 & 0.016 & 0.432 & 0.452 & 0.464 & 0.474 & 0.494 \\
Claude 4.5 Sonnet (big5) & Balanced Accuracy & 0.307 & 0.018 & 0.274 & 0.295 & 0.307 & 0.319 & 0.344 \\
Claude 4.5 Sonnet (big5) & Macro-F1 & 0.250 & 0.021 & 0.212 & 0.235 & 0.250 & 0.265 & 0.290 \\
Grok 4 (big5) & Accuracy & 0.454 & 0.018 & 0.421 & 0.442 & 0.454 & 0.467 & 0.490 \\
Grok 4 (big5) & Balanced Accuracy & 0.289 & 0.016 & 0.259 & 0.278 & 0.289 & 0.300 & 0.322 \\
Grok 4 (big5) & Macro-F1 & 0.227 & 0.020 & 0.191 & 0.214 & 0.227 & 0.240 & 0.266 \\
GPT-5 Mini (big5) & Accuracy & 0.434 & 0.018 & 0.401 & 0.421 & 0.434 & 0.446 & 0.470 \\
GPT-5 Mini (big5) & Balanced Accuracy & 0.273 & 0.016 & 0.243 & 0.262 & 0.273 & 0.284 & 0.306 \\
GPT-5 Mini (big5) & Macro-F1 & 0.199 & 0.018 & 0.167 & 0.187 & 0.199 & 0.211 & 0.236 \\
GPT-4o Mini (big5) & Accuracy & 0.431 & 0.018 & 0.397 & 0.418 & 0.431 & 0.443 & 0.467 \\
GPT-4o Mini (big5) & Balanced Accuracy & 0.270 & 0.016 & 0.241 & 0.259 & 0.270 & 0.281 & 0.302 \\
GPT-4o Mini (big5) & Macro-F1 & 0.195 & 0.018 & 0.164 & 0.183 & 0.195 & 0.207 & 0.231 \\
DeepSeek V3 (big5) & Accuracy & 0.440 & 0.017 & 0.408 & 0.428 & 0.440 & 0.452 & 0.474 \\
DeepSeek V3 (big5) & Balanced Accuracy & 0.279 & 0.016 & 0.250 & 0.268 & 0.279 & 0.290 & 0.312 \\
DeepSeek V3 (big5) & Macro-F1 & 0.206 & 0.018 & 0.175 & 0.194 & 0.206 & 0.218 & 0.242 \\
\hline
\end{tabular}
\end{table}

\begin{table}[t]
\centering
\caption{Figure 2 results. Performance as a function of trait dimensionality. Mean performance with 95\% participant-level bootstrap confidence intervals.}
\label{tab:fig2_results}
\small
\begin{tabular}{llllllllll}
\hline
\textbf{Model} & \textbf{Traits} & \textbf{Metric} & \textbf{Mean} & \textbf{Std} & \textbf{2.5\%} & \textbf{25\%} & \textbf{50\%} & \textbf{75\%} & \textbf{97.5\%} \\
\hline
LBM-2 & 74 & Accuracy & 0.628 & 0.019 & 0.592 & 0.615 & 0.627 & 0.640 & 0.668 \\
LBM-2 & 74 & Macro-F1 & 0.478 & 0.026 & 0.428 & 0.460 & 0.478 & 0.496 & 0.531 \\
Llama 3.1 8B & 74 & Accuracy & 0.427 & 0.018 & 0.394 & 0.415 & 0.427 & 0.439 & 0.462 \\
Llama 3.1 8B & 74 & Macro-F1 & 0.164 & 0.009 & 0.147 & 0.158 & 0.164 & 0.170 & 0.181 \\
Claude 4.5 Sonnet & 74 & Accuracy & 0.468 & 0.016 & 0.436 & 0.457 & 0.468 & 0.479 & 0.500 \\
Claude 4.5 Sonnet & 74 & Macro-F1 & 0.245 & 0.015 & 0.216 & 0.235 & 0.245 & 0.255 & 0.273 \\
Grok 4 & 74 & Accuracy & 0.379 & 0.020 & 0.339 & 0.366 & 0.379 & 0.392 & 0.416 \\
Grok 4 & 74 & Macro-F1 & 0.180 & 0.011 & 0.158 & 0.173 & 0.181 & 0.188 & 0.204 \\
\hline
LBM-2 & 40 & Accuracy & 0.615 & 0.019 & 0.578 & 0.601 & 0.614 & 0.628 & 0.650 \\
LBM-2 & 40 & Macro-F1 & 0.443 & 0.024 & 0.399 & 0.426 & 0.443 & 0.460 & 0.491 \\
Llama 3.1 8B & 40 & Accuracy & 0.428 & 0.018 & 0.394 & 0.415 & 0.428 & 0.440 & 0.462 \\
Llama 3.1 8B & 40 & Macro-F1 & 0.164 & 0.009 & 0.146 & 0.158 & 0.164 & 0.170 & 0.182 \\
Claude 4.5 Sonnet & 40 & Accuracy & 0.472 & 0.018 & 0.437 & 0.460 & 0.472 & 0.484 & 0.507 \\
Claude 4.5 Sonnet & 40 & Macro-F1 & 0.257 & 0.017 & 0.225 & 0.245 & 0.256 & 0.267 & 0.291 \\
Grok 4 & 40 & Accuracy & 0.399 & 0.018 & 0.364 & 0.386 & 0.399 & 0.410 & 0.433 \\
Grok 4 & 40 & Macro-F1 & 0.194 & 0.012 & 0.169 & 0.185 & 0.194 & 0.202 & 0.218 \\
\hline
LBM-2 & 20 & Accuracy & 0.622 & 0.019 & 0.587 & 0.610 & 0.622 & 0.634 & 0.657 \\
LBM-2 & 20 & Macro-F1 & 0.455 & 0.025 & 0.406 & 0.437 & 0.456 & 0.471 & 0.503 \\
Llama 3.1 8B & 20 & Accuracy & 0.427 & 0.018 & 0.393 & 0.415 & 0.426 & 0.439 & 0.462 \\
Llama 3.1 8B & 20 & Macro-F1 & 0.164 & 0.009 & 0.148 & 0.158 & 0.164 & 0.170 & 0.181 \\
Claude 4.5 Sonnet & 20 & Accuracy & 0.460 & 0.017 & 0.428 & 0.448 & 0.460 & 0.472 & 0.494 \\
Claude 4.5 Sonnet & 20 & Macro-F1 & 0.243 & 0.017 & 0.210 & 0.232 & 0.243 & 0.255 & 0.276 \\
Grok 4 & 20 & Accuracy & 0.384 & 0.019 & 0.347 & 0.371 & 0.383 & 0.396 & 0.420 \\
Grok 4 & 20 & Macro-F1 & 0.184 & 0.011 & 0.163 & 0.177 & 0.184 & 0.191 & 0.205 \\
\hline
LBM-2 & 10 & Accuracy & 0.564 & 0.019 & 0.525 & 0.551 & 0.565 & 0.577 & 0.602 \\
LBM-2 & 10 & Macro-F1 & 0.386 & 0.026 & 0.334 & 0.369 & 0.387 & 0.404 & 0.436 \\
Llama 3.1 8B & 10 & Accuracy & 0.424 & 0.017 & 0.391 & 0.413 & 0.423 & 0.435 & 0.459 \\
Llama 3.1 8B & 10 & Macro-F1 & 0.163 & 0.009 & 0.146 & 0.157 & 0.163 & 0.169 & 0.180 \\
Claude 4.5 Sonnet & 10 & Accuracy & 0.468 & 0.018 & 0.435 & 0.455 & 0.468 & 0.480 & 0.503 \\
Claude 4.5 Sonnet & 10 & Macro-F1 & 0.252 & 0.017 & 0.219 & 0.240 & 0.252 & 0.264 & 0.287 \\
Grok 4 & 10 & Accuracy & 0.384 & 0.019 & 0.345 & 0.372 & 0.384 & 0.397 & 0.423 \\
Grok 4 & 10 & Macro-F1 & 0.172 & 0.010 & 0.153 & 0.165 & 0.172 & 0.179 & 0.192 \\
\hline
LBM-2 & 5 & Accuracy & 0.480 & 0.018 & 0.445 & 0.468 & 0.480 & 0.491 & 0.514 \\
LBM-2 & 5 & Macro-F1 & 0.256 & 0.020 & 0.217 & 0.242 & 0.256 & 0.271 & 0.296 \\
Llama 3.1 8B & 5 & Accuracy & 0.427 & 0.018 & 0.393 & 0.414 & 0.427 & 0.438 & 0.464 \\
Llama 3.1 8B & 5 & Macro-F1 & 0.164 & 0.009 & 0.147 & 0.158 & 0.164 & 0.170 & 0.181 \\
Claude 4.5 Sonnet & 5 & Accuracy & 0.464 & 0.017 & 0.429 & 0.453 & 0.465 & 0.476 & 0.496 \\
Claude 4.5 Sonnet & 5 & Macro-F1 & 0.250 & 0.017 & 0.217 & 0.239 & 0.251 & 0.263 & 0.283 \\
Grok 4 & 5 & Accuracy & 0.370 & 0.018 & 0.334 & 0.358 & 0.369 & 0.382 & 0.407 \\
Grok 4 & 5 & Macro-F1 & 0.167 & 0.009 & 0.149 & 0.161 & 0.167 & 0.174 & 0.186 \\
\hline
\end{tabular}
\end{table}

\end{document}